\documentclass[11pt]{article}
\usepackage{coling2020}
\usepackage{times}
\usepackage{latexsym}
\usepackage{coling-nert}
\usepackage{natbib}
\usepackage{url}

\usepackage{microtype}
\usepackage{amsmath}
\usepackage{subcaption}

\usepackage{graphicx}
\usepackage{xcolor}
\usepackage{cleveref}

\colingfinalcopy

\definecolor{ss}{HTML}{22AA22}
\renewcommand{\ss}[1]{\textsc{\textcolor{ss}{#1}}}
\newcommand{\ssblue}[1]{\textsc{\textcolor{blue}{#1}}}
\newcommand{\ssbad}[1]{\textsc{\textcolor{red}{#1}}}
\newcommand{\pr}[2]{{#1\ensuremath{_\ss{#2}}}}
\newcommand{\prblue}[2]{{#1\ensuremath{_\ssblue{#2}}}}

\title{Supersense and Sensibility:\\ Proxy Tasks for Semantic Annotation of Prepositions}
\author{Luke Gessler \quad Shira Wein \quad Nathan Schneider \\ 
  Georgetown University \\
  {\{\emldisplay{lg876@georgetown.edu}{lg876}, \emldisplay{sw1158@georgetown.edu}{sw1158},
  \emldisplay{nathan.schneider@georgetown.edu}{nathan.schneider}\}\texttt{@georgetown.edu} }}

\date{}

\makeatletter
\renewcommand{\paragraph}{%
  \@startsection{paragraph}{4}%
  {\z@}{0.60ex \@plus 1ex \@minus .2ex}{-1em}%
  {\normalfont\normalsize\bfseries}%
}
\makeatother
\begin{document}
\maketitle

\begin{abstract}
Prepositional supersense annotation is time-consuming and requires expert training. Here, we present two sensible methods for obtaining prepositional supersense annotations \emph{indirectly} by eliciting surface substitution and similarity judgments. Four pilot studies suggest that both methods have potential for producing prepositional supersense annotations that are comparable in quality to expert annotations.
\end{abstract}


\section{Introduction}

\blfootnote{
    \hspace{-0.65cm}  
    This work is licensed under a Creative Commons 
    Attribution 4.0 International License.
    Licence details:
    \url{http://creativecommons.org/licenses/by/4.0/}.
}

Prepositions are highly ambiguous function words which can express a wide variety of relationships \citep{litkowski-hargraves-2005, tratz_semantically-enriched_2011}. 
Supersenses have been proposed as an analytic framework for studying their lexical semantics, but extant gold-annotated corpora \citep[e.g.][]{schneider_comprehensive_2018,peng-etal-2020-corpus} are small because preposition supersense annotation is a relatively complex annotation task that requires substantial training and time.

We ask whether preposition supersense annotation could be made cheaper and quicker with crowdsourced labor. This will require ``sensible'' annotation tasks accessible to non-experts. In this work we present two possible designs for \textbf{proxy} tasks for crowdsourcing from which supersense labels can be recovered indirectly. These designs involve in-context \emph{substitution} and \emph{similarity} judgments. Based on four in-house pilot experiments, we conclude that both designs are promising as methods for building a large preposition supersense--annotated corpus, 
and that they differ in how difficult they are for the workers and for the researchers.
%
However, the setup should be considered a proof of concept, as we work with an idealized pool of workers (in-house computational linguistics graduate students).
Future research will be necessary to ascertain whether the paradigm is suited to na\"{i}ve crowdworkers as well.

\section{Preposition Supersenses}
\label{sec:ps}

Prepositions\footnote{And adpositions, more generally---but since all data in the present work is from English, we will write {\it preposition} throughout.} can express many different kinds of semantic relations. 
\citet{schneider_comprehensive_2018} present SNACS, a coarse-grained annotation framework for prepositions encoding these relations.   
The meanings of prepositions are expressed in terms of supersenses, of which there are 50 in SNACS v2.5.\footnote{See: \url{http://flat.nert.georgetown.edu/supersenses/}}
For instance, the preposition \textit{in} can be used to express time, place, and other relations: ``I rented an apartment \pr{in}{Locus} Boston'', ``I hope to see you \pr{in}{Time} the future''.%
\footnote{%
In SNACS, prepositions are actually annotated with {\it two} supersenses: one for their {\it scene role}, which describes the ``basic semantic relation between the preposition-linked elements'', and one for their {\it function}, which captures the ``semantic relation literally or metaphorically present in the scene [...] highlighted by the choice of adposition''. 
Whenever the scene role does not match the function role, the two are notationally separated with a pipe, as in ``You are meticulous in your work and it shows \pr{in}{Manner|Locus} my smile.''
For the purposes of the present work, we will simplify our discussion of the prepositional supersense tagging task and speak of it as if it consisted of assigning a single label (the scene role and function tags, concatenated).
}

Compared to other tagging tasks, supersense annotation is relatively hard: the SNACS guidelines, which are only for prepositional supersenses, are around 100 pages in length, and a single preposition can often have multiple plausible annotations which must be carefully considered before a final decision.


\section{Two Task Designs}

Our ultimate goal is to obtain supersense labels for prepositions in context from crowdsourced data.
One possible technique would be to provide definitions and canonical examples of each label, or subsets of labels, and ask the crowdworker which label most closely applies to the annotation target \citep{munro_crowdsourcing_2010}. 
Another tactic would be to decompose our labels into more readily intuitive semantic \emph{features} \citep{reisinger_semantic_2015}.
But given the extensive semantic range of the many prepositions we seek to annotate, both of these approaches seem difficult to achieve with crowdworkers.

Instead, we explore what we term \emph{proxy tasks}:\footnote{This is unrelated to the term ``proxy task'' as used by \citet{mostafazadeh-etal-2016-story}, where it is used to refer to intrinsic evaluations for word embeddings.} rather than teach and elicit supersenses (or semantic features associated with supersenses) directly, we elicit judgments of surface substitution\slash similarity, as has been done by previous work on word sense crowdsourcing (\cref{sec:related}).
This approach leverages current annotated data in combination with the proxy annotations to infer supersense labels via crowdsourcing.
Here we outline two different approaches for framing a crowdsourcing task from which annotations can be derived. 
Details will be explained in more depth when we turn to discuss our pilot studies.

\subsection{Preposition Substitution}
\label{sec:parsub}

This design consists of two crowdsourced tasks and requires an unlabeled corpus $\mathcal{U}$.  
First, in the \emph{generation~task}, we identify an unlabeled instance $\langle s, t\rangle \in \mathcal{U}$, where $s$ is a sentence and $t \in s$ is the \emph{target preposition} to be disambiguated.
The sentence $s$ is presented to a crowdworker, and the worker is asked to provide a substitute $t^\prime$ for $t$ which approximately preserves the meaning of $s$ when substituted with $t$  and does not contain $t$.
E.g., for the sentence ``The book is {\bf by} the lamp'', ``close to'' and ``near'' would both be good substitutes because ``The book is {\bf close to} the lamp'' and ``The book is {\bf near} the lamp'' both have similar meanings. 
Substitutes can be anything: they do not have to be prepositions, and they do not have to be a single word.
By the end of this task, several potential substitutes $t^\prime_1, \ldots, t^\prime_n$ will have been proposed by workers, but this data alone is not enough to infer a supersense label.

More information is collected in the second task, the \emph{selection task}. 
The substitutes from the generation task $t^\prime_1,\ldots,t^\prime_n$ populate a multiple-choice list, and crowdworkers choose all items on the list which are acceptable substitutes for $t$ in $s$. 
Once enough crowdworkers have completed the selection task, we are left with a frequency distribution over the substitutes. (For an example of such a distribution, see \cref{fig:radar}.) 

These distributions must somehow be turned into supersense labels. 
One way to do this is to source labeled instances $\langle s, t, \ell_\text{gold}\rangle$ for the tasks above from a gold-labeled corpus $\mathcal{L}$. 
This would allow a classifier to predict each instance's annotation from its substitution distribution, which could then be used to label unseen data. 
However, one concern is that a statistical classifier based on substitutes would be no more accurate for infrequent prepositions than training a supersense classifier directly; a set of heuristic rules for disambiguating supersenses that could use the selected substitutes may be effective here.


\subsection{Neighbor Selection}


This design consists of a single crowdsourced task and requires a labeled corpus $\mathcal{L}$, an unlabeled input corpus $\mathcal{U}$, and some similarity function $\mathit{sim}(x,y)$ that can compare two unlabeled instances $\langle s_1, t_1 \rangle$, $\langle s_2, t_2\rangle$ and represent as a real number how similar the two usages of prepositions $t_1$ and $t_2$ are in their contexts.%
\footnote{We deliberately do not mention a specific metric or representation here, since there are many ways to implement this design. As we describe in \cref{sec:p3}, we use cosine distance between supersense membership softmax vectors from a supersense tagger for our pilots in this work, though one could imagine other implementations, such as Euclidean distance between raw or fine-tuned BERT embeddings.}

An unlabeled instance $\langle s,t \rangle \in \mathcal{U}$ is selected, which we call the \textit{target} instance. 
$\mathit{sim}$ is used to compare it to every instance in $\mathcal{L}$, and the top $k$ most similar inst/ances in $\mathcal{L}$ are retrieved with their labels, $\langle s_1, t_1, \ell_1 \rangle, \ldots, \langle s_k, t_k, \ell_k \rangle$. We call these retrieved instances the target's \textit{neighbors}. Neighbors may optionally be filtered, e.g.~to ensure that no label $\ell$ is represented more than once among $\ell_1, \ldots, \ell_k$.

The target sentence $s$ is presented to crowdworkers along with $s_1, \ldots, s_k$ from the neighbors, with the target preposition indicated in each, and crowdworkers are asked to select any neighbors for which the usage of the preposition $t_i$ in $s_i$ most resembles the usage of $t$ in $s$.%
\footnote{Strictly speaking, this design could be implemented with or without any formal guidance given to workers on what should count as ``similar enough'' for this task, but for our pilot studies, we deliberately choose not to give workers any guidance. Our motivation for this was to see what kind of granularity we would get from worker judgments without any explicit instruction.}
For example, if the target sentence were ``I was booked {\bf at} the hotel'', ``There was no cabbage {\bf at} the store'' would be a neighbor to choose, while ``My technician arrived {\bf at} 11 pm'' would not be a good neighbor to choose. The predicted supersense tag is taken from the neighbor sentence that was selected most often by crowdworkers.

\subsection{Comparison}

We hypothesize that the substitution design, a two-task design with a non-trivial annotation inference step, would be more time- and resource-intensive than the one-task neighbor selection design. On the other hand, we expect the neighbor selection design to work well only as long as $\mathcal{L}$ is big enough to contain relevant neighbors, which may not be the case for rarer supersense tags.
Moreover, the neighbor selection design relies on a similarity metric which may not always successfully find good neighbors.
Thus it is worth exploring whether the two designs have complementary strengths.

\section{Pilot Studies}

In order to assess the quality and characteristics of these two designs before large-scale deployment, we conducted a series of pilot studies with a small number of participants. Participants in each study were drawn from a group of several graduate students, all of whom had at least some familiarity with SNACS. This is an unrealistic quality for crowdworkers to have, but our aim is to assess how well these designs can work in ideal conditions.

\subsection{Pilot 1: Substitute Generation}

First, we carry out the generation task of the substitution design. Five common prepositions are selected for annotation: {\it for}, {\it with}, {\it to}, {\it from}, {\it in}. For each preposition, 30 instances were retrieved from STREUSLE, with 10--20 supersenses represented across all of a given preposition's instances. Seven workers were shown the instances and asked to write a single substitution per instance according to the guidelines in \cref{sec:parsub}.

We find that for any given instance, all workers tended to produce different substitutes. If workers had tended to converge on a single substitute, there might have been some hope of recovering a supersense label directly, but as there was almost never consensus, an additional task is needed to determine tags.

\begin{figure}[t!]
\centering 
\begin{subfigure}[t]{0.47\textwidth}
    \centering
    \includegraphics[width=\linewidth]{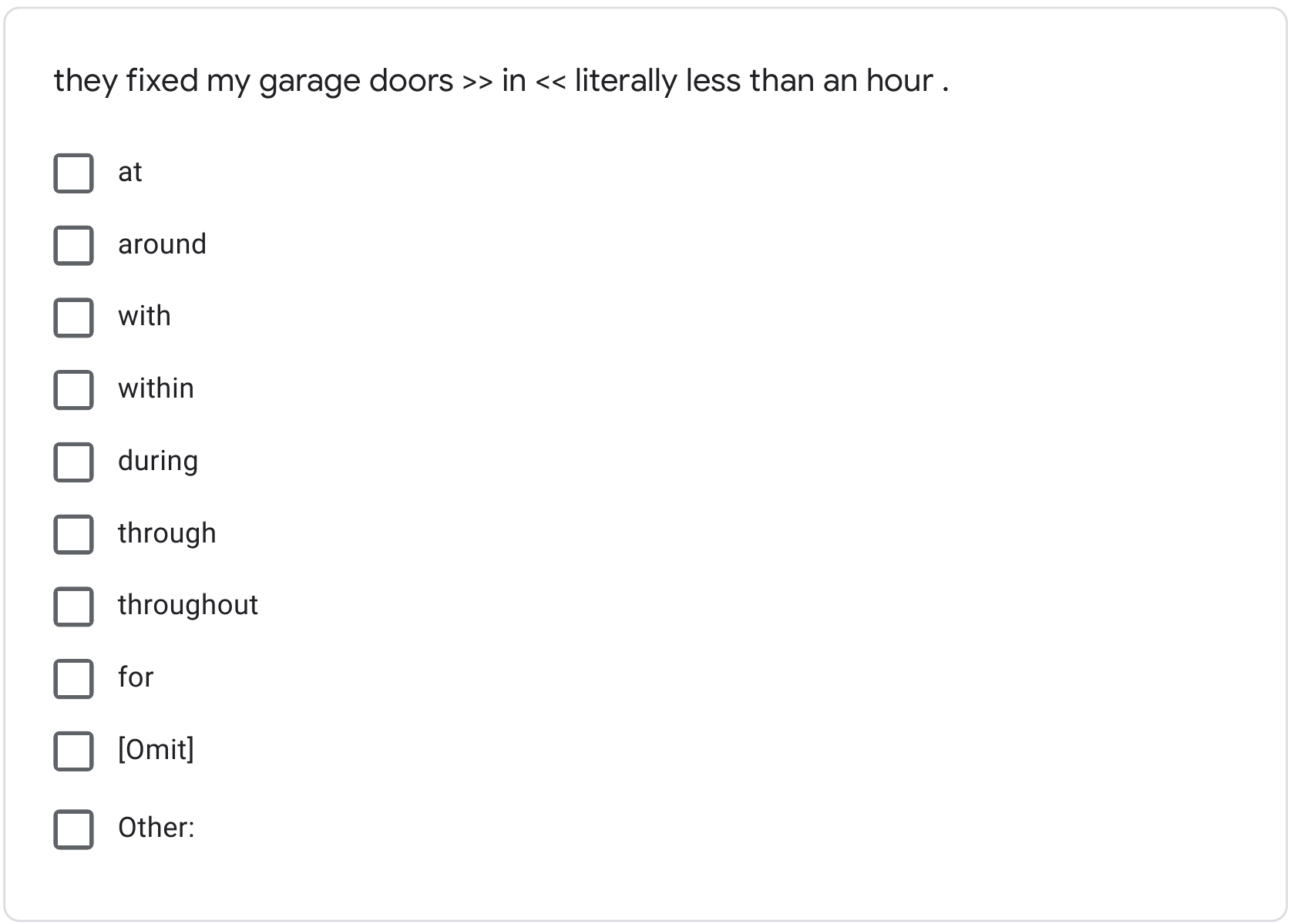}
    \caption{A prompt from pilot 2, substitute selection, for a single instance: the target preposition is indicated with angle brackets, and workers are tasked with selecting substitutes which roughly preserve the sentence's meaning.}
    \label{fig:pilot2}
\end{subfigure}%
\hspace{0.02\textwidth}
\begin{subfigure}[t]{0.47\textwidth}
    \centering
    \includegraphics[width=\linewidth]{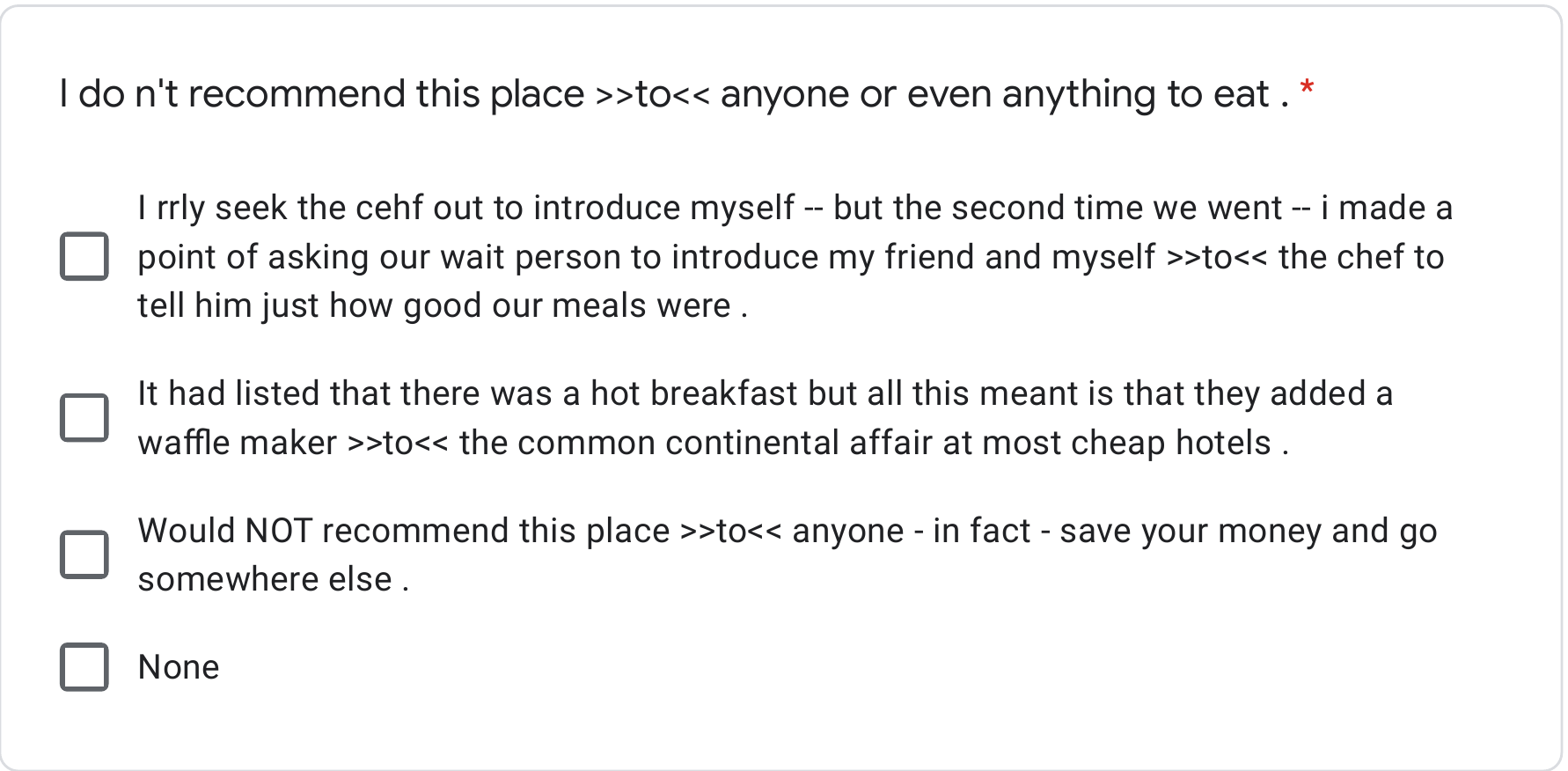}
    \caption{A prompt from pilot 3. Although six retrieval strategies were used to contribute neighbors for this candidate sentence, only three unique neighbors were retrieved across them.}
    \label{fig:p3}
\end{subfigure}
\caption{Prompts for pilots 2 and 3.}
\end{figure}

\subsection{Pilot 2: Substitute Selection}
\begin{figure}[t!]
\centering 
\begin{subfigure}[t]{0.47\textwidth}
    \centering
    \includegraphics[height=1.7in]{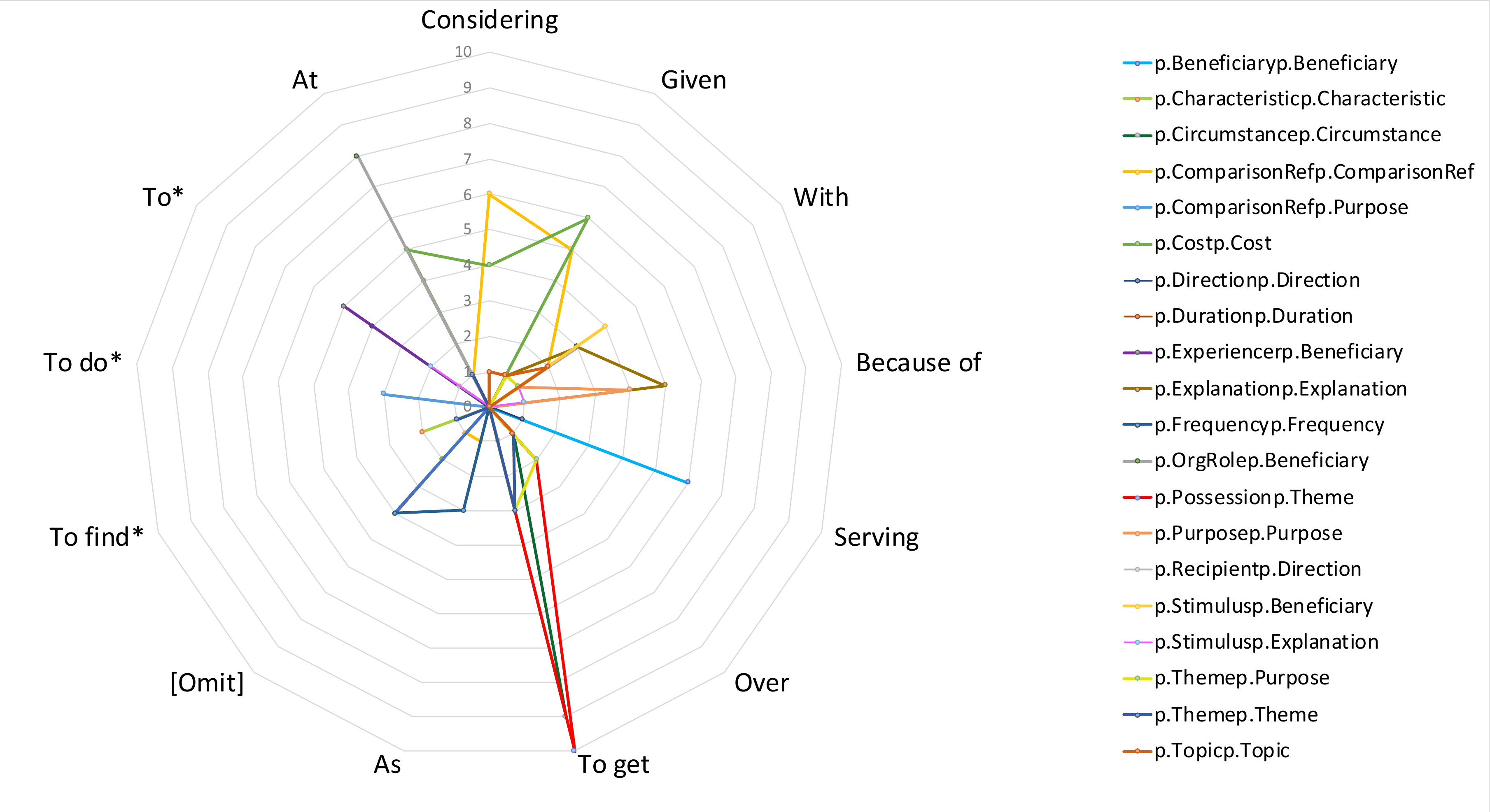}
    \caption{ Substitutes for ``for'' from pilot 2.  }
\end{subfigure}
\hspace{0.02\textwidth}
\begin{subfigure}[t]{0.47\textwidth}
    \centering
    \includegraphics[height=1.7in]{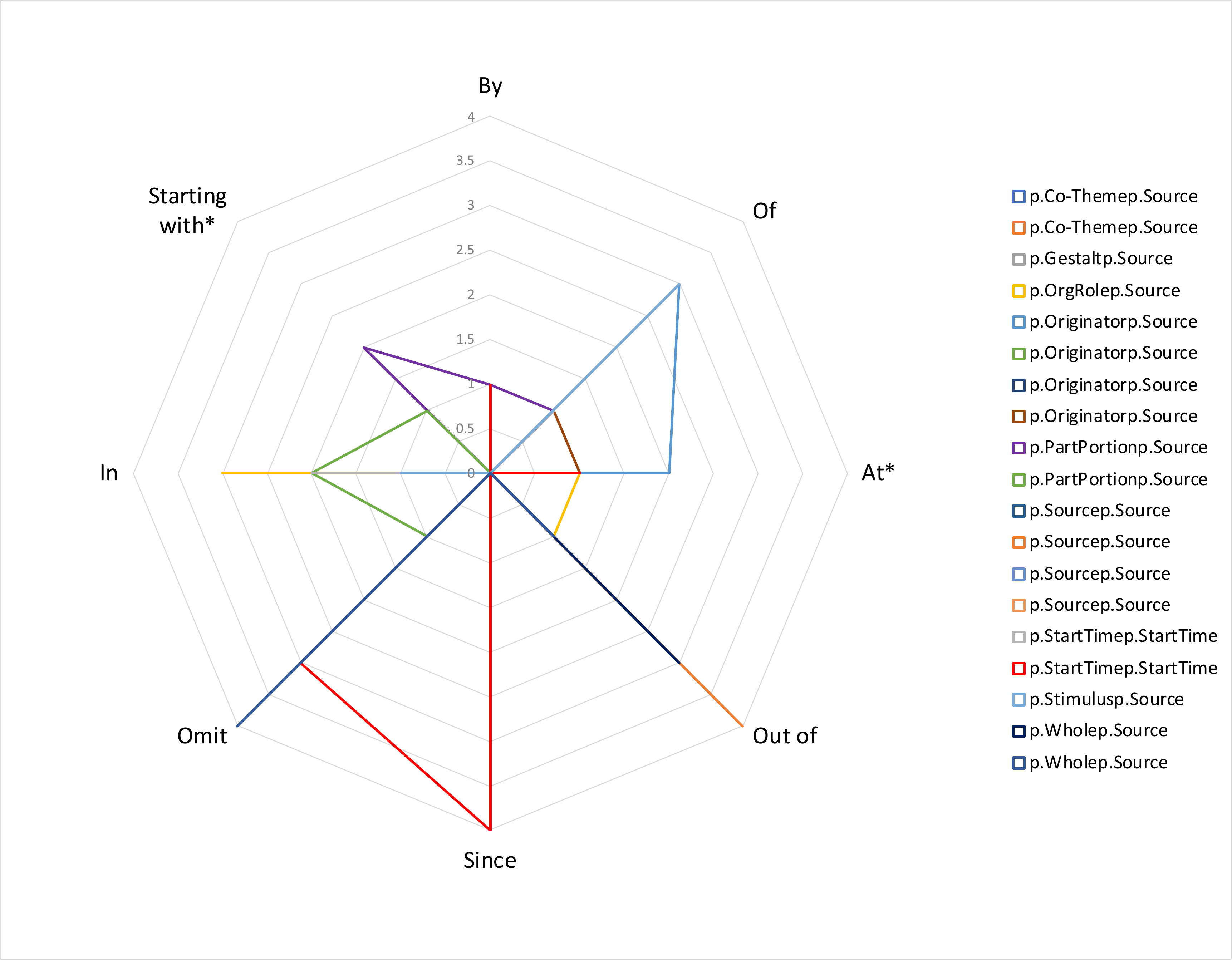}
    \caption{ Substitutes for ``from'' from pilot 2.  }
\end{subfigure}
\hspace{0.02\textwidth}

\vspace{1.0em}

\begin{subfigure}[t]{0.47\textwidth}
    \centering
    \includegraphics[height=1.7in]{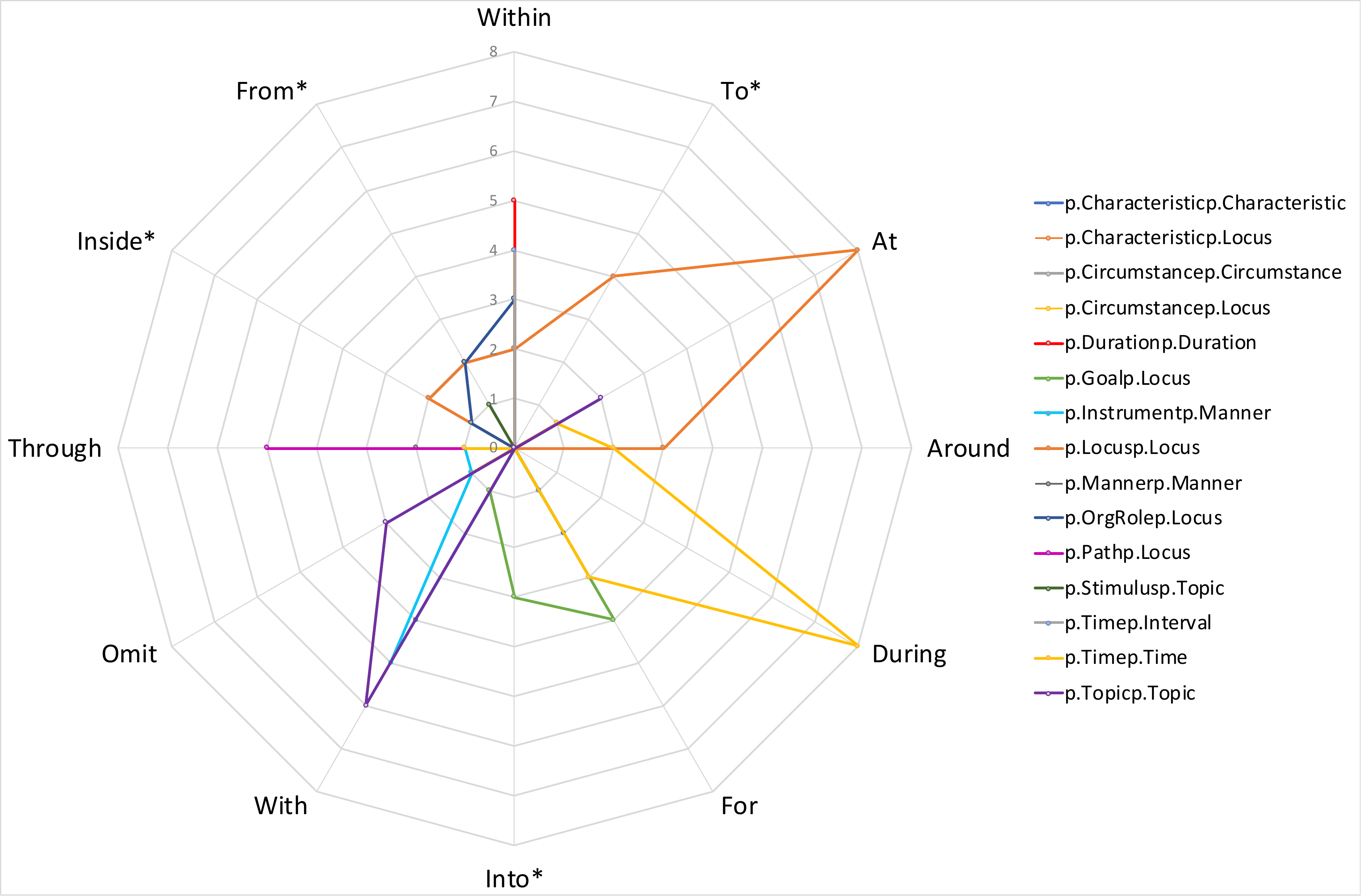}
    \caption{ Substitutes for ``in'' from pilot 2.  }
\end{subfigure}
\hspace{0.02\textwidth}
\begin{subfigure}[t]{0.47\textwidth}
    \centering
    \includegraphics[height=1.7in]{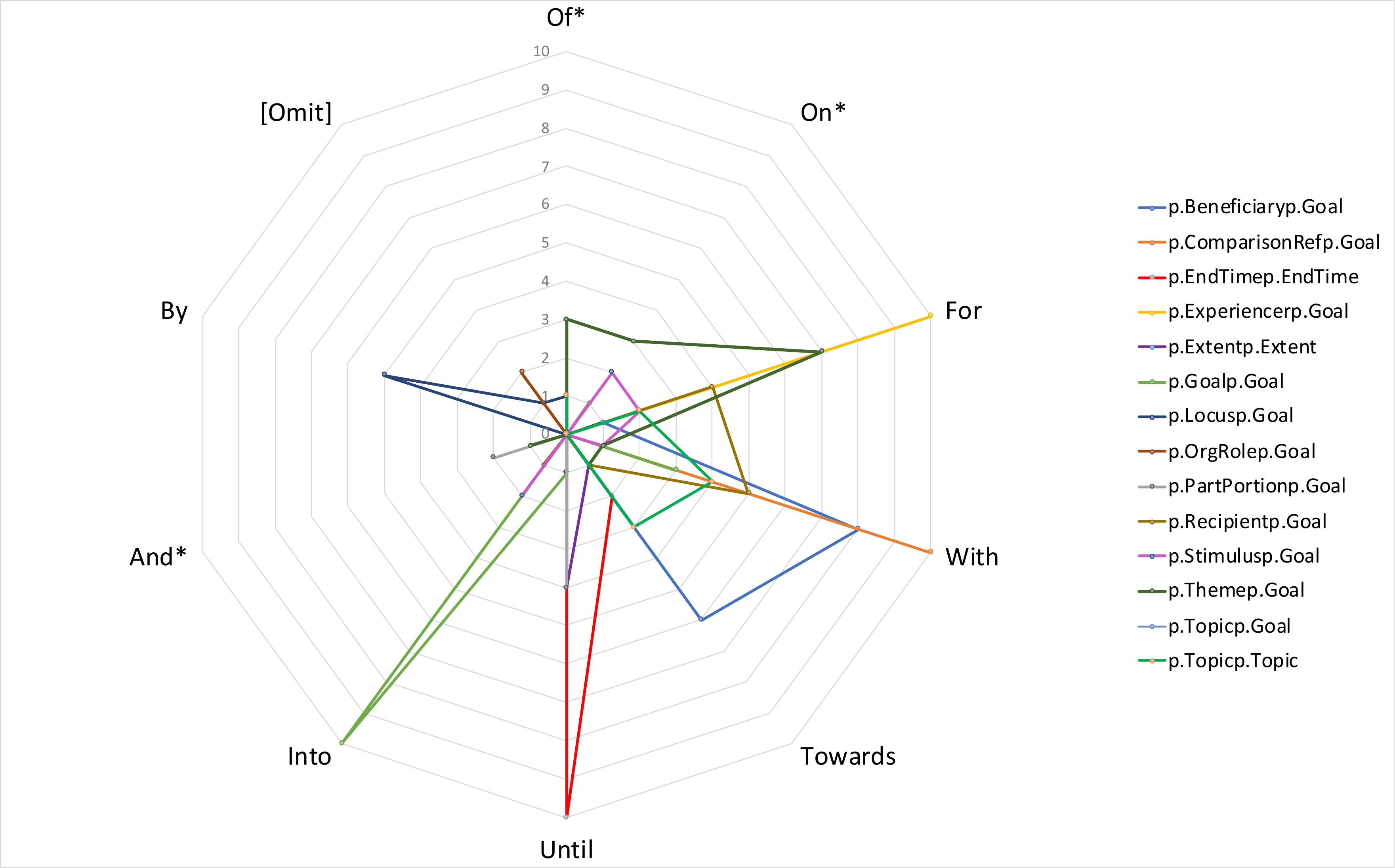}
    \caption{ Substitutes for ``to'' from pilot 2.  }
\end{subfigure}
\hspace{0.02\textwidth}

\vspace{1.0em}

\begin{subfigure}[t]{0.47\textwidth}
    \centering
    \includegraphics[height=1.7in]{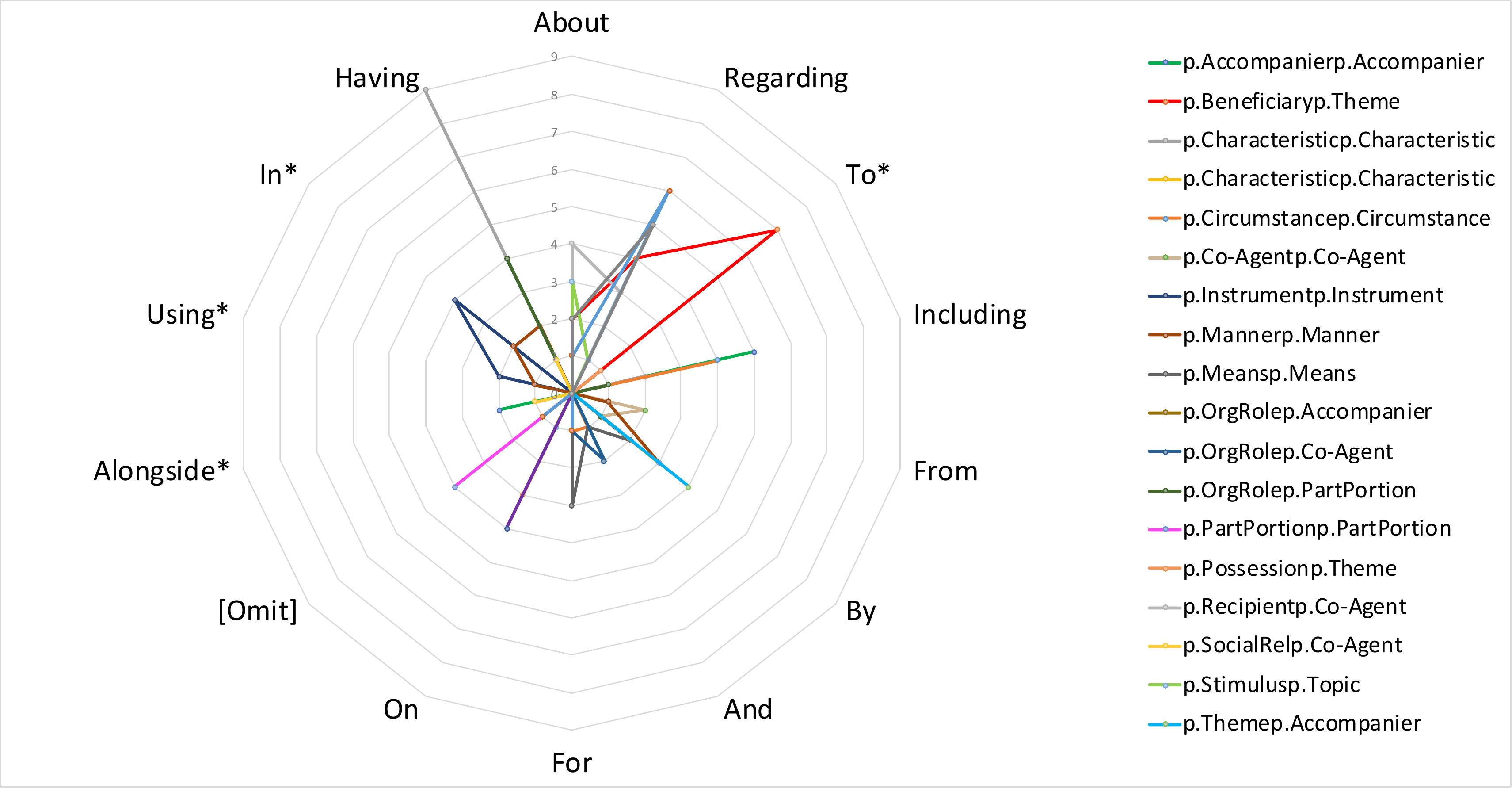}
    \caption{ Substitutes for ``with'' from pilot 2.  }
\end{subfigure}
\hspace{0.02\textwidth}
\caption{Substitutes from pilot 2. 
Each ``spoke'' represents a substitute, and every point on each colored line represents the frequency of a substitute used for an instance that was gold-labeled with a particular supersense tag. 
For instance, in (c), we can see that of all the instances of ``in'' that were gold-tagged with \ss{Goal|Locus}, {\it with} was picked once, {\it into} three times, and {\it for} four times. 
Only substitutes that were picked three or more times across all 5 prepositions are shown. Asterisked paraphrases were written in by workers during the selection task and were not produced by the generation task.}
\label{fig:radar}
\end{figure}

In pilot study 2, we carry out the second task of the substitution design, substitute selection. For each of the 5 prepositions investigated in pilot 1, we select 30 instances from STREUSLE, yielding a total of 150 instances for this pilot. The substitutes for each preposition are selected by taking the top 8 most commonly provided substitutes for all instances of that preposition in pilot 1. In addition to these substitutes, workers could also indicate that none of the 8 substitutes are appropriate, either by choosing ``[Omit]'', or providing an alternative substitute. (See  \cref{fig:pilot2}.) Seven workers participated in this pilot.

As described in \cref{sec:parsub}, in a full implementation of this task, the next step would be to train classifiers to predict the tag for each instance from its substitute distribution. However, our dataset is too small to train a classifier, so we inspect our results qualitatively. 

Recall that for a single instance $\langle s, t \rangle$, the substitute selection task leaves us with a frequency distribution over $t$'s substitutes $t_1^\prime, \ldots, t_k^\prime$. If we aggregate these distributions over every instance of $t$, and then group them by the supersense label $\ell$, we are left with distributions as shown in \cref{fig:radar}, each of which tells us the number of times a particular substitute was chosen across all instances of $t$ that had the tag $\ell$. While no firm conclusions can be drawn because of the limited size of the data, we see 
enough separation between regions to suggest that supersenses have distinguishable substitute distributions---and therefore, the distributions obtained from crowdworkers would be effective for classifying the supersense.

\subsection{Pilot 3: Neighbor Selection Retrieval Strategy}
\label{sec:p3}

The neighbor selection design retrieves similar sentences from a labeled corpus for a similarity judgment.
This design critically relies on the quality of the top $k$ instances which are retrieved. In pilot 3 we compare several strategies for retrieving the top $k$ instances. 

Our $\mathit{sim}(\langle s_1, t_1 \rangle,\langle s_2, t_2\rangle)$ relies on 
\citeposs{liu_lexical_2020}
supersense tagger, and is implemented as the cosine similarity between the tagger's vectors of supersense label probabilities for $t_1$ and $t_2$ as they are used in $s_1$ and $s_2$. 
In order to obtain high-quality vectors for the entire corpus, we use a strategy similar to jackknife resampling: we partition the corpus's documents so that they are approximately balanced by token count, yielding 5 splits $\mathcal{L}_1, \ldots, \mathcal{L}_5$. Now, for each $\mathcal{L}_i$, the vectors for instances contained within $\mathcal{L}_i$ are obtained by training the tagger on the other 4 splits and predicting on $\mathcal{L}_i$.

For each of the same 5 prepositions as above, we first identify from the STREUSLE test and development sets 8 instances: 4 of which the tagger correctly predicts, and 4 of which it incorrectly predicts. We focus on incorrect predictions especially because it is important to understand how well this method works when the tagger's predictions are wrong.
(There would be little purpose to crowdsourcing if it were only accurate for instances that the tagger already classified correctly!)

Next, six retrieval strategies are identified. These strategies differ in three parameters: their ranking method (cosine ranking, versus random ranking as a baseline); whether neighbors are required to feature the same preposition as the target (the same-word constraint); and whether neighbors are required to be tagged with the same supersense as the target (the same-supersense constraint; this is an oracle of sorts to potentially reduce situations where none of the options are relevant). This yields 8 possibilities, but we exclude random ranking without the same-word constraint because it would obviously yield bad results.

\begin{table}
\centering 
    \begin{subtable}[t]{0.48\textwidth}
        \centering
        \small
        \begin{tabular}{l@{~~}c@{~~}c}
            Strategy & Votes & Majority \\\hline
            None & \hphantom{0}21 & \hphantom{0}6 \\
            Random, same-word, same-supersense & \hphantom{0}22 & \hphantom{0}8\\
            Random, same-word & \hphantom{0}10 & \hphantom{0}2 \\
            Cosine, same-word, same-supersense & \hphantom{0}74 & 24 \\
            Cosine, same-word & \hphantom{0}79 & 27\\
            Cosine, same-supersense & \hphantom{0}74 & 24\\
            Cosine, no constraints & \hphantom{0}79 & 27\\\hline
            Theoretical Maximum & 120 & 40\\
        \end{tabular}
        \caption{A tabulation of the number of times a strategy's retrieved neighbor was selected by a worker in pilot 3. Three workers examined 40~instances, so the maximum possible tally is 120, shown in the ``votes'' column. The ``Majority'' column tallies the neighbors that received a majority among the workers (ties possible).}\label{tab:p3}
    \end{subtable}
    \hspace{0.02\textwidth}
    \begin{subtable}[t]{0.48\textwidth}
        \centering
        \small
        \begin{tabular}{l@{~~}c@{~~}c@{~~}c}
            Case & Tagger & Crowd & ``None'' \\\hline
            1 (Tagger correct, gold present) & 17/17 & 17/17 & \hphantom{0}0/17\\
            2 (Tagger incorrect, gold present) & \hphantom{0}0/12 & \hphantom{0}6/12 & \hphantom{0}5/12 \\
            3 (Tagger correct, gold absent) & \hphantom{0}3/3\hphantom{0} & \hphantom{0}0/3\hphantom{0} & \hphantom{0}2/3\hphantom{0} \\
            4 (Tagger incorrect, gold absent) & \hphantom{0}0/8\hphantom{0} & \hphantom{0}0/8\hphantom{0} & \hphantom{0}5/8\hphantom{0} \\
        \end{tabular}
        \caption{Tagger and crowd accuracy for pilot 4 grouped by whether the tagger correctly predicted the target's gold tag and whether the target's gold tag was present among the 5 neighbors. ``None'' column indicates how many times ``None'' was chosen by the crowd. (These results should not be taken to be indicative of real-world performance, since pilot 4's dataset was deliberately constructed to include correctly and incorrectly tagged instances in equal proportions, while in a real scenario we expect the tagger would be correct more often than not.)}
        \label{tab:p4}
    \end{subtable}
\caption{Data from pilots 3 and 4.}
\end{table}

For each of the 40 instances, each strategy contributes a single neighbor, and these neighbors are deduplicated before being presented to a worker so they could choose the best neighbors (possibly multiple if there is a tie), as shown in \cref{fig:p3}. A ``None'' option is again provided in case no neighbor is close enough to the target. In order to determine the success of each strategy, we simply tally the number of times a neighbor that was contributed by that strategy was selected by a worker, of which there were three for this study. The results are given in  \cref{tab:p3}. The results show that, on the whole, unconstrained cosine ranking successfully finds neighbors that workers deem relevant. 
Moreover, overall, strategies with the same-word and same-supersense constraints do not outperform cosine without constraints.
That the same-supersense methods, which have access to the gold label for the target, do not outperform the unconstrained cosine strategy suggests that the latter is reasonably robust on its own. It remains possible that these strategies could have complementary strengths, though our data here is too limited to speculate.

\subsection{Pilot 4: Neighbor Selection}

In pilot 4, we carry out a small-scale version of the neighbor selection design, using the same 40 instances from pilot 3 and using cosine ranking to identify 5 neighbors for each instance. We do not use the same-word or same-supersense constraints from pilot 3, but we do introduce a diversity constraint, which requires that all neighbors have different supersense tags. This 
maximizes our odds that one of the 5 neighbors will have the correct tag for the target.

We analyze results from 5 workers, who were shown each target with its 5 neighbors and asked to choose the most relevant neighbors, or ``None'', similar to \cref{fig:p3}. 
We use the workers' plurality vote to select a neighbor, and use the neighbor's gold tag as the target's predicted tag.  
Ideally, our participants will arrive on a plurality vote for the correct neighbor when the gold tag is present among them, and if the gold tag is not present, they will either select ``None'' or choose neighbors with gold tags that are similar, but not identical, to the target's gold tag. 

Our results show that our participants performed fairly close to this ideal: broadly, whenever the crowd selects a neighbor it usually has the correct gold label, and when no neighbor is a good choice or there is something pathological about the instance, the crowd often selects ``None''. 
In our discussion we will partition this pilot's results according to two parameters: whether the tagger correctly predicted the target preposition's supersense tag, and whether a neighbor was retrieved whose gold tag matched the target preposition's tag (thereby affording workers the opportunity to correctly label the instance). A summary of results is given in \cref{tab:p4}. 

\begin{figure}
    \centering
    \begin{subfigure}[b]{0.48\textwidth}
        \small
        \noindent \textbf{Target sentence:} One time we even left after sitting at the table for 20 minutes and not being greeted with$_{\ssblue{Means},\ssbad{Possession|Accompanier}}$ a drink order .
        
        \vspace{0.4em}
            
        \noindent  \textbf{Neighbor sentences:}
        
        \noindent (1) I can not tell you how often I am complimented \prblue{on}{Topic} my hair ( style AND color ) !
        
        \noindent (2) A simple follow - up phone call \prblue{with}{Co-Agent} a woman quickly turned into a nightmare .
        
        \noindent (3) wow , the representative went way above and beyond in helping me \prblue{with}{Theme} my account set up .
        
        \noindent ({\bf 4, selected}) This store is proof that you can fool people \prblue{with}{Means} good advertising .
        
        \noindent (5) She has taken care of my sweet girl for almost 4 years now and I would not let Gracee go \prblue{with}{Accompanier|Co-Agent} anyone besides her !!!
        
        \noindent (6) None
        \vspace{5\baselineskip} 
        \vspace{1em}
        \caption{A case 2 instance from pilot 4 where workers identified the neighbor with the gold label.}\label{fig:greeted}
    \end{subfigure}
    \hspace{0.02\textwidth}
    \begin{subfigure}[b]{0.48\textwidth}
        \small
        \noindent \textbf{Target sentence:} The atmosphere is your typical indie outfit with old movie posters and memorabilia from$_{\ssblue{StartTime},\ssbad{Originator|Source}}$ the 70's and 80's .
        
        \vspace{0.4em}
            
        \noindent \textbf{Neighbor sentences:}
        
        \noindent (1) I just got back \prblue{from}{Source} france yesterday and just missed the food already !
        
        \noindent (2) prepared the road test with a driving ... prepared the road test with a driving school in edmonton , but my instructor only trained me in a narrow street , hence I took one 90 minute lesson \prblue{from}{OrgRole|Source} the Noble driving school to learn the skill of changing lane , and found them very friendly and professional .
        
        \noindent (3) yet again it was a great stay \prblue{from}{StartTime} begiinning to end .
        
        \noindent (4) We order take out \prblue{from}{Originator|Source} here all the time and we are never disappointed .
        
        \noindent (5) \# 2 the decor is tasteful and artistic , \prblue{from}{PartPortion|Source} the comfortable chairs to the elegant light fixtures .... and ( most importantly ) \# 3 the food is FANTASTIC .
        
        \noindent ({\bf 6, selected}) None
        \vspace{1em}
        \caption{A case 2 instance from pilot 4, where workers chose ``None'' even though a gold-tagged neighbor was present.}\label{fig:starttime}
    \end{subfigure}    
    \caption{Two instances from Pilot 4. Tags from the gold-annotated source corpus are given in blue, and the supersense tagger's predictions (which in these two examples are also incorrect) are given in red.}
    \label{fig:p4}
\end{figure}

\paragraph{Case 1: tagger correct, gold-tagged neighbor present} When the tagger produces the correct supersense label for the target preposition and also succeeds in finding a gold-tagged neighbor, the crowd always chooses the correct neighbor. This is an encouraging result for the question of whether crowdworkers can perform worse than the tagger: it seems that in cases where the tagger finds it easy to tag correctly and produce a gold-tagged neighbor, humans are also able to easily recognize the right answer.

\paragraph{Case 2: tagger incorrect, gold-tagged neighbor present} The case when the tagger was incorrect but still manages to retrieve a relevant neighbor is an important one, because it is where humans have the opportunity to \emph{improve on} the tagger's performance, rather than keep up with it (case 1) or fall behind it (cases 3 and 4). In case 2, workers managed to retrieve the correct tag 6$/$12 times and in the remaining cases unambiguously choose ``None'' or tie vote for ``None'' 5$/$6 times. This result tells us that in case 2, workers are able to either choose the neighbor with the correct tag or refuse to choose a neighbor most of the time. A representative example is given in \cref{fig:greeted}: here, the gold label of ``with'' in the target sentence is \ss{\small Means}, while the model gives the similar but mistaken tag \ss{\small Possession|Accompanier}.\footnote{This is because ``with a drink order'' is describing the manner of the greeting, not an item that was in someone's possession for the main event, as in ``I arrived \pr{with}{Possession|Accompanier} my box to ship.''} 
	                       
A more interesting instance is given in \cref{fig:starttime}. The tagger has mistagged the target instance, ``\pr{from}{StartTime} the 70's and 80's'', and although it still manages to find a gold neighbor ``\pr{from}{StartTime} beginning to end'', the crowd rejects it in favor of ``None''. 
Although SNACS assigns \ss{\small StartTime} to both instances, this obscures a notable difference in meaning: the former use describes when an object was made, and the latter use the beginning of an event.
The most likely explanation, then, is that the crowd perceived this difference in meaning and decided the latter instance was not similar enough in meaning to be acceptable.
This instance therefore constitutes evidence that workers are capable of making some distinctions that are \emph{more} nuanced than those made by SNACS.

The remaining instances from case 2 mostly either demonstrated the tagger's misunderstanding of subtle distinctions which humans were able to recover from (either by choosing ``None'' or the gold neighbor), although some other instances featured metaphor which posed a challenge for both the tagger and humans. 
The instance ``Food is awful and the place caters \pr{to}{Beneficiary|Goal} the yuppy crowd .'' has the expression \emph{caters to} which has a literal sense (`serving food to') that differs from its metaphorical sense (`pandering to').
Interpreting a metaphorical expression literally or metaphorically will almost always entail a difference in supersense and is an issue for SNACS in general.


\paragraph{Case 3: tagger correct, no gold-tagged neighbor} The 3 times where the tagger correctly predicted the target's gold tag but did not manage to retrieve any neighbors with this tag were all somehow exceptional.
Looking into them, we determined that in one case, the target was not well treated by the SNACS guidelines and could plausibly have been annotated differently, and that the two other instances were metaphorical and similarly could have been annotated differently (cf.~case 2).
While it is difficult to generalize from so few instances, we see that the tagger's top predicted supersense is sometimes not represented in the cosine-retrieved neighbors, even though the cosine measure makes use of the tagging model.
This suggests that it might be advantageous in future iterations of this design to consult the tagger's prediction and require at least one neighbor to have the same supersense.

\paragraph{Case 4: tagger incorrect, no gold-tagged neighbor} In the last case where the tagger was wrong and no neighbors with the correct tag were retrieved, we found similarly hard cases  which were due to vagueness in the SNACS guidelines, well-formedness issues, or other relatively uncommon 
causes. 
In these cases, we felt that expert human annotators also would have struggled to choose the correct tag or could have defensibly argued for different analyses. 
One instance, for example, contains a crucial typo: ``it was a little to high dollar \pr{for}{} me''. 
The ``to'' should have been a ``too'', and ``too'' is highly connected to the fact that this instance of ``for'' is involved in a comparison.

                       

\paragraph{Summary} We have seen in pilot 4 that human crowdworkers, in aggregate, are generally cautious, good at being confident when they should be, and choosing ``None'' when no neighbors are appropriate.%
\footnote{The extent to which this is also true of \emph{individual} crowdworkers before their responses have been aggregated is unclear from the work we have discussed here.}
It should be noted that the ``None'' result is not simply a dead end---if an instance receives a ``None'', additional steps can in principle be taken to attempt to elicit an answer, e.g.~by fetching a new batch of neighbors and putting it back into the annotation pool. 
We also reiterate that the dataset in pilot 4 was deliberately constructed so that half of its instances would have incorrect tag predictions. 
Even still, if we take pilot 4 results as a measure of performance, our crowdsourcing method delivers higher-quality annotations than the tagger alone (which can already achieve $F_1$ in the low 80s on STREUSLE's test split), demonstrating the potential of this approach. 

\section{Related Work}\label{sec:related}

Crowdsourcing on Amazon Mechanical Turk has been a popular method for scaling up linguistic annotation. 
Early studies on its efficacy for semantic annotations like word sense disambiguation, textual entailment, and word similarity \citep{snow_cheap_2008} and psycholinguistic studies and judgment elicitation \citep{munro_crowdsourcing_2010} have shown that crowdsourced annotations can be as good as if not even better than annotations produced using traditional methods. 
Several studies have focused specifically on word sense annotations for content-words like nouns, adjectives, and verbs \citep{rumshisky_crowdsourcing_2011,jurgens_embracing_2013,biemann_crowdsourcing_2010,biemann_turk_2012,tsvetkov-etal-2014-augmenting}, but function words have received less attention. 
To our knowledge, the only work carried out specifically on prepositional sense annotation using a non-traditional annotation methodology is due to \citet[\S 4.2]{tratz_semantically-enriched_2011}, who describes a process by which existing prepositional sense annotations were refined by three annotators, two of which have unspecified levels of linguistic or other competencies. 
We conjecture that one reason why content-words have been favored over function-words for crowdsourced annotation is that their comparatively less abstract meanings make reasoning about their semantics more approachable to linguistically na\"ive crowdworkers, simplifying task design, while for function words, it can be difficult to tap into crowdworkers' intuitions without changing the task considerably. 

Some work has investigated gamification with the hope that bringing gamelike elements would allow crowdworkers to produce good annotations without a traditional training process, in some cases achieving performance on par with or better than expert annotation \citep{fort_rigor_2020,hartshorne_verbcorner_2014,schneider_comprehensive_2014}.  
Independent from the matter of whether to crowdsource or gamify, some have modified their annotation schemes with an eye explicitly to annotation expense (in terms of time or money). 
In the QA-SRL annotation scheme proposed by \citet{he_question-answer_2015}, plain-language questions are used to describe the predicate-argument structures of verbs instead of formalisms such as frames or predicates, rendering it theory- and formalism-neutral and easier to explain to non-expert workers.  


Our work is different from the work we have reviewed here in that we have attempted to have participants solve a task that is \emph{not} the same task we would have given to an expert annotator. 
Pursuing such a proxy task, as we have termed it, introduces the challenge of turning proxy data into gold data, but reduces the need for worker training. 
Proxy tasks have been successfully pursued in other domains, like in the ESP game designed by \citet{von_ahn_labeling_2004} for image labeling, where player data for a game played with an image---the proxy task---is used to infer image labels.

\section{Conclusion}
We have presented two designs for deriving prepositional supersense tags from crowdsourced tasks, and we have investigated their efficacy through four pilot studies, finding that both hold promise for producing high-quality prepositional supersense annotations. We have seen that the two designs differ in their complexity and performance characteristics: the neighbor selection design, while consisting of only one task instead of two, requires a gold-annotated corpus of sufficient size to give every tag sufficient coverage, while the substitution design could reveal, bottom-up, clusters of usages that may not be well-represented in the training data.

We have made several idealizations throughout this work: all data was drawn from STREUSLE, guaranteeing that it would be homogeneous with respect to genre, and crowdworkers had some knowledge of the SNACS guidelines which likely made them better at the tasks than real-world crowdworkers. Moreover, we studied only 5 common prepositions covering 20 or so supersenses out of SNACS's 50. In future work, we intend to implement these designs on platforms such as Amazon Mechanical Turk to further investigate these designs' efficacy and the extent to which these idealizations affect our results.

\section*{Acknowledgements}
This research was supported in part by NSF award IIS-1812778. We thank our pilot participants Shabnam Behzad, Michael Kranzlein, Yang Liu, Emma Manning, and Jakob Prange, and we also thank Yang (again) and Siyao Peng for their helpful comments on a draft of this paper. Finally, we thank our anonymous reviewers from LAW XIV and DMR 2020 for their detailed and thoughtful comments.

\bibliographystyle{acl_natbib}
\bibliography{ourbib}

\end{document}